\documentclass[review]{elsarticle}

\usepackage{blindtext}
\usepackage{graphicx}
\usepackage{multirow}
\usepackage{booktabs}
\usepackage{comment}
\usepackage{xcolor}
\usepackage{amssymb}
\usepackage{mathtools}
\usepackage{hyperref}
\usepackage{tikz}
\usepackage{xcolor,colortbl}
\usepackage{url}

\begin{document}
\begin{frontmatter}
\title{GR-RNN: Global-Context Residual Recurrent  Neural Networks for Writer Identification}

\author[author1]{Sheng He\corref{cors}}
\cortext[cors]{Corresponding author}
\ead{heshengxgd@gmail.com}

\author[author2]{Lambert Schomaker}
\ead{l.r.b.schomaker@rug.nl}

\address[author1]{Department of Radiology, Boston Children's Hospital, Harvard Medical School, Harvard University, Boston, MA 02215, USA}
\address[author2]{Bernoulli Institute for Mathematics, Computer Science and Artificial Intelligence, University of Groningen, 
	PO Box 407, 9700 AK, Groningen, The Netherlands}

\begin{abstract}
This paper presents an end-to-end neural network system to identify writers through handwritten word images, which jointly integrates global-context information and a sequence of local fragment-based features.
The global-context information is extracted from the tail of the neural network by a global average pooling step. The sequence of local and fragment-based features is extracted from a low-level deep feature map which contains subtle information about the handwriting style.
The spatial relationship between the sequence of fragments is modeled by the recurrent neural network (RNN) to strengthen the discriminative ability of the local fragment features.
We leverage the complementary information between the global-context and local fragments, resulting in the proposed global-context residual recurrent neural network (GR-RNN) method.
The proposed method is evaluated on four public data sets and experimental results demonstrate that it can provide state-of-the-art performance.
In addition, the neural networks trained on gray-scale images provide better results than neural networks trained on binarized and contour images, indicating that texture information plays an important role for writer identification.

The source code will be available: 

\url{https://github.com/shengfly/writer-identification}.
\end{abstract}


\end{frontmatter}

\section{Introduction}

Forensic writer identification refers to the task of identifying a specific writer of a piece of handwriting, which has potential applications in forensic document examination~\cite{pervouchine2007extraction} and historical manuscript analysis~\cite{cilia2020minimum,brink2012writer,dhali2020feature}.
The classical methods~\cite{schomaker2004automatic,bulacu2007text,christlein2017writer,chahi2020local} use shape or texture features of handwritten text to recognize the writer, which requires a large amount of image information per sample in order to obtain a statistically reliable feature vector~\cite{cilia2020minimum,bulacu2007text}.
Therefore, most studies focus on writer identification using page-level document images which contain several paragraphs or sentences.
High recognition rates are reported on public datasets, such as 98.5\%~\cite{wu2014offline,lai2020encoding}, 99.7\%~\cite{tang2016text} and 98.0\%~\cite{khan2018dissimilarity} on the IAM~\cite{marti2002iam}, CVL~\cite{kleber2013cvl} and Firemaker~\cite{schomaker2000forensic} datasets, respectively.
However, writer identification on small amount of data such as a word image is still a challenging problem, due to the lack of information for modeling the handwriting style of the writer. The handcrafted features completely fail in word-based writer identification, resulting in low recognition rates of 37.2\% and 30.0\% on the IAM and CVL datasets, respectively~\cite{he2019deep}.

Convolutional neural networks (CNN)~\cite{lecun1998gradient,krizhevsky2012imagenet} can extract semantic and deep abstract features by stacking several convolutional layers. This method has been widely used in pattern recognition, including writer identification~\cite{tang2016text,fiel2015writer,keglevic2018learning,chen2019semi}.
Previous studies~\cite{nguyen2019text,he2019deep,chen2019semi,he2020fragnet} show that the deep features extracted from the trained neural networks provide a good result in writer identification, even on the handwritten word images.
A typical neural network~\cite{krizhevsky2012imagenet} consists of several convolutional layers which generate feature maps by linear convolutional filters followed by a global average pooling (GAP~\cite{lin2013network}).
The GAP which sums out over the spatial information captures the global context while decreasing the granularity of features~\cite{bai2020deep}. While the ensuing generalized representation is useful in general object classification, it misses the local discriminative details which describe subtle handwriting style information. 
The FragNet~\cite{he2020fragnet} approach extracts local detailed information on fragments sampled from input images and deep feature pyramids, which can enrich the generated feature and describe the handwriting style in detail.
However, in the FragNet approach, the spatial context between fragments is ignored.

In fact, the global-context information and the features of local fragments are complementary to each other. Integrating those two types (i.e., levels) of information can improve the performance of writer identification based on very small amount of handwritten text.
In this paper, we propose to compute the handwriting-style information from word images, using the sequence of local fragments $x_i$ which are extracted from CNN feature maps to model the handwritten word images. 
To capture the context within the fragment sequence, we employ the gated recurrent unit (GRU)~\cite{cho2014learning}, which is a gating mechanism in recurrent neural networks (RNN), $f^i=\text{GRU}(x_i,f^{i-1})$, where $f^0$ is usually initialized to zero, to enhance the discriminative capacity of local fragment $f^i$. 
Inspired by the residual networks~\cite{he2016deep,ilyes2018residual}, we use the residual RNN, which is defined as $f^i=\text{GRU}(x_i,f^{i-1})+x_i$. Thus, the feature $f^i$ of local fragment contains not only the between-fragment context information in the sequence of the fragments that are passing through the RNN, but also the original fragment feature $x_i$.
In addition, instead of initializing the $f^0$ to zero, we set the $f^0$ to the global-context obtained by the global average pooling of the last layer on the neural network.
Thus, the proposed method is named as global-context residual recurrent neural network (GR-RNN for short), which integrates information from the global-context and the local sequence of fragments for writer identification.

The main contribution of this paper is fourfold: 1) We propose to consider the word image as a sequence of fragments segmented on the deep feature map and apply a recurrent neural network to extract the contextual information between fragments;
2) Two kinds of complementary information from the global-context and the local fragments are integrated by the proposed global-context residual recurrent neural network (GR-RNN);
3) The proposed method is evaluated for writer identification based on word images on four public datasets and it provides better results than state-of-the-art methods;
4) Different types of images are evaluated for writer identification and we found that the performance on the gray-scale word images is better than the performance on the corresponding binarized and contour word images, indicating that the CNN focuses on the textural information on ink traces, e.g., caused by the writing instrument and varying pressure. 

This paper is organized as follows: a brief summary of related work is presented in Section~\ref{sec:relatedwork}. The detailed description of the proposed GR-RNN is given in Section~\ref{sec:method} and the experimental results is shown in Section~\ref{sec:exps}. We give a conclusion in the last section.

\section{Related Work}
\label{sec:relatedwork}

This section presents a brief summary of classical methods for writer identification and the discussed approaches are not all inclusive. 
A comprehensive survey of writer identification can be found in~\cite{darganwriter} for interested readers.
We divide methods into two groups: handcrafted features and deep learning.

\subsection{Handcrafted features for writer identification}
The study of writer identification can be traced back to the earliest work of Arazi~\cite{arazi1977handwriting} in 1977, who used the histogram of the run-lengths of the background intensity as feature vectors for writer identification.
After that, many different handcrafted features have been proposed for solving the forensic writer identification problem.
Generally, there are three basic principles to design effective handcrafted features for writer identification.

\textit{Principle 1}: designing textural-based features by the joint feature distribution~\cite{he2017beyond}. 
The Hinge feature~\cite{schomaker2004automatic,bulacu2007text} is the joint distribution of the two angles along the handwritten contours of the ink traces.
Brink et al.~\cite{brink2012writer} found that the ink width takes an important role for writer identification in historical documents wrote by the Quill instrument, resulting in the Quill feature which is the joint distribution of ink direction and ink width.
The oriented Basic Image Features (oBIFs)~\cite{newell2014writer} which is the joint distribution of six Derivative-of-Gaussian filters at different scales is used for writer identification, with a delta encoding method. 
The curvature-free COLD feature~\cite{he2017writer}, which is the joint distribution of the length and orientation of line fragments, is proposed for writer identification, combining the run-length of general patterns.
More features generated by the joint feature distribution, such as CoHinge and QuadHinge, can be found in~\cite{he2017beyond}.

\textit{Principle 2}: encoding the grapheme-based features can improve the performance.
In~\cite{wu2014offline}, the SIFT descriptors with the corresponding scales and orientations are used to describe word images and a codebook is trained using the self-organizing map~\cite{schomaker2004automatic}. 
Christlein et al.~\cite{christlein2017writer} used RootSIFT descriptors and GMM supervectors as encoding method to describe the characteristic handwriting of an individual writer. 
Abdi et al.~\cite{abdi2015model} proposed a method to synthesize graphemes based on the beta-elliptic model for Arabic writer identification.
Inspired by Fraglets~\cite{bulacu2007text}, Khalifa et al.~\cite{khalifa2015off} segmented connected components based on contours as graphemes which are normalized for building the codebook and the global feature descriptor. 
Based on the fact that junctions are prevalent in handwritten scripts, the junclets~\cite{he2015junction}, which is the distribution of line width from junction points to boundary in different angles, has been used for writer identification based on a codebook trained from a set of detected junctions in handwritten images.
Recently, a log path signature, called pathlets, is proposed in~\cite{lai2019offline,lai2020encoding}, which is defined as a consecutive segment on the polygonized contour and described by iterated integrals.

\textit{Principle 3}: combining the textural and grapheme-based features can improve the performance.
Schomaker and Bulacu~\cite{schomaker2004automatic} proposed to combine edge-Hinge features and the connected-component contours (CO$^3$) on uppercase Western scripts. 
Later, these methods are extended to Fraglets and contour-Hinge~\cite{bulacu2007text} and their combination provides better performance than all individual features involved in the combination.
Siddiqi and Vincent~\cite{siddiqi2010text} proposed an automatic writer identification method by combining two types of features: the codebook-based feature captures the frequent small writing fragments of each writer and the textural feature captures the visual attributes of writing.
Ahmad et al.~\cite{khan2018dissimilarity} combined scale-invariant feature transform (SIFT) and RootSIFT descriptors in a set of Gaussian Mixture Models (GMM). 
Lai et al.~\cite{lai2020encoding} applied a combination of pathlet and unidirectional SIFT features for writer identification on historical document images.
Alaa et al.~\cite{sulaiman2019length} used a combination of the deep and handcrafted descriptors extracted on a patch with the size of 30$\times$30 from the handwritten images.

\subsection{Deep learning for writer identification}
Deep learning has also been widely used for writer identification.
Fiel and Sablatnig~\cite{fiel2015writer} used the CaffeNet, which contains five convolutional layers and three fully-connected layers for deep feature extraction.
A two-stream neural network with shared weights is used in~\cite{xing2016deepwriter} for writer identification based on line images in English and isolated characters in Chinese. 
An unsupervised deep feature learning method is proposed in~\cite{christlein2017unsupervised} and the neural network is trained using the pseudo-labels generated by the cluster indices of the clustered SIFT descriptors extracted on 32$\times$32 image patches. 
In~\cite{keglevic2018learning}, a triplet network is used to compute a similarity measure for image patches and the deep learned features are encoded using the vector of locally aggregated descriptors (VLAD) to generate the global feature vector for each document image.
The conditional AutoEncoder is used in~\cite{hosoe2018offline} to extract deep features.
In~\cite{chen2019semi}, a semi-supervised feature learning method is proposed to learn discriminative representation  and the learned deep features are encoded using VLAD for offline writer identification based on word-based images.
In~\cite{nguyen2019text}, local deep features are extracted using convolutional neural networks in character images and their sub-regions and deep features of images from the tuples are aggregated to form global features for text-independent writer identification.
The AlexNet architecture with transferred knowledge from ImageNet is employed in~\cite{rehman2019automatic} to extract discriminating visual features from multiple representations of image patches for English and Arabic writer identification.
Cilia et al.~\cite{cilia2020end} evaluated several state-of-the-art deep architectures for writer identification in medieval manuscripts with a relatively small training dataset.
Javidi and Jampour~\cite{javidi2020deep} proposed an end-to-end framework by conjugating deep learned features and a traditional handwriting descriptor for writer identification.
All of these methods extract local features using deep learning and aggregate the extracted local features to form a feature vector for computing the handwriting style similarity between different individuals. 

Our previous works in~\cite{he2019deep,he2020fragnet} train the convolutional neural networks to predict the writer identity directly from the word images.
In~\cite{he2019deep}, a multi-task learning strategy is applied to train the neural network with the writer identification as the main task. 
Deep features learned from the auxiliary tasks are adapted to the main task to improve the performance.
The  FragNet has been proposed in~\cite{he2020fragnet}, which is the neural network trained on fragments extracted from both input image and the feature maps of another neural network trained for extracting feature pyramid.

\subsection{Residual learning}
A residual network~\cite{he2016deep} can be briefly described as:
$y = F(x,w) + x$
where $y$ is the output, $x$ is the input, $w$ is the parameter and $F$ is the neural network.
With the shortcut path, the network only needs to learn the residual information $F(x,w)=y-x$.
The network structure $F$ can be convolutional layers~\cite{he2016deep}, recurrent neural networks~\cite{ilyes2018residual} or self-attention block in transformer~\cite{vaswani2017attention}.
In our proposed method, we use the recurrent neural network (RNN) to capture the spatial context information between fragments.
Unlike traditional RNN using zero values as the initial state, we use the global-context information as the initial state to integrate the global and local information.

\section{Proposed method}
\label{sec:method}

In this section, we present the details of the proposed method for writer identification.
The proposed method focuses on integrating the global-context information from the global average pooling and the sequence of fragment-based local representations segmented on deep learned feature maps.

\subsection{The main architecture}

\begin{figure}
    \centering
    \includegraphics[width=0.8\textwidth]{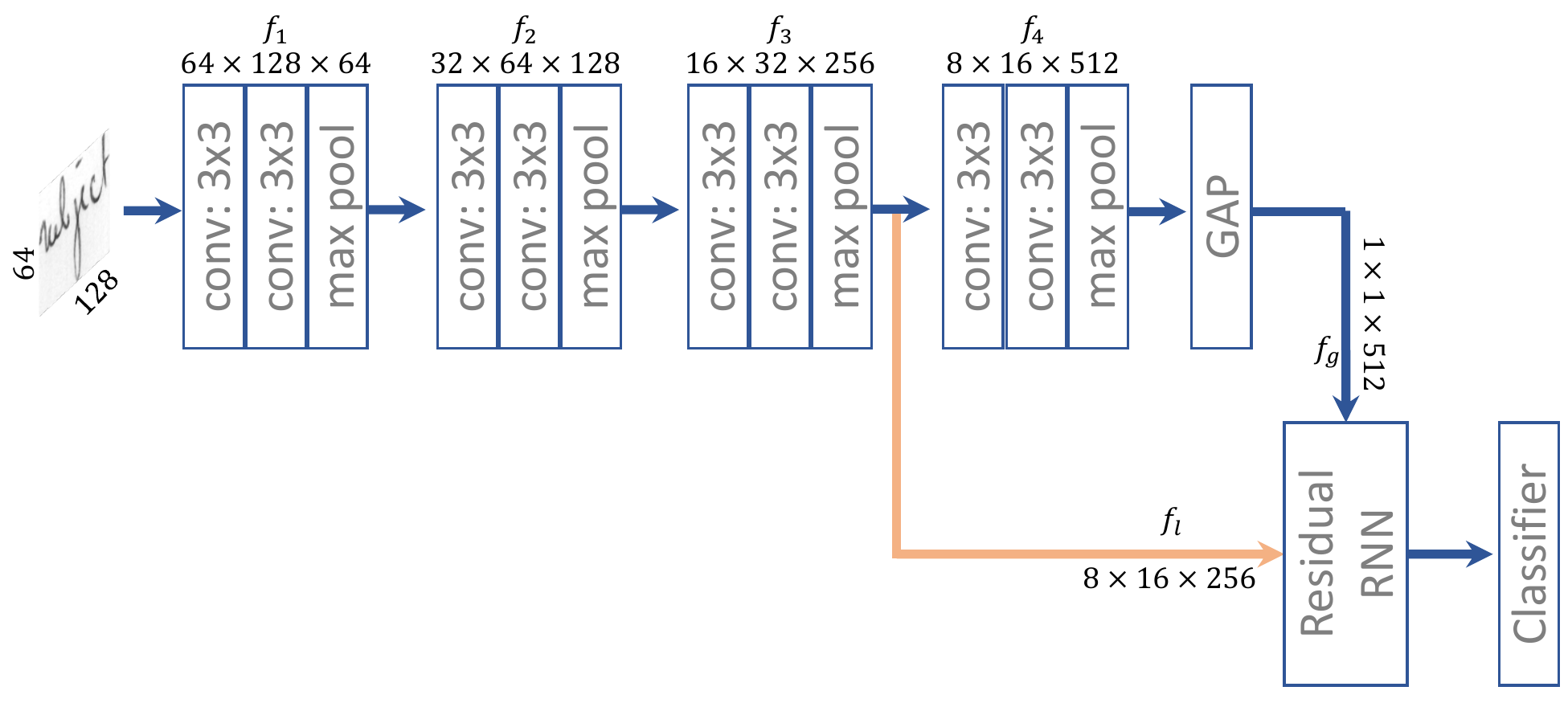}
    \caption{Illustration of the main structure of neural network. It contains four blocks $f_1,f_2,f_3,f_4$,  Each block contains two convolutional layers and one max-pooling layer. $f_g$ is the global-context extracted by the global average pooling (GAP) and $f_l$ is the local feature tensor extracted after the third block $f_3$.
    The size format of each tensor is denoted by height$\times$width$\times$depth.}
    \label{fig:framework}
\end{figure}

As shown in Fig.~\ref{fig:framework}, the network consists of four blocks and each block has two convolutional layers and one max-pooling layer.
The size of the input image is 64$\times$128$\times$1 and the size of the feature maps $f_i$ on each block is denoted as $h\times w\times c$ where $h$, $w$ and $c$ are the height, width and depth (the number of channels), respectively. 
The parameters of all convolutional layers are the same: kernel size is fixed to 3$\times$3, the size of stride and padding is fixed to 1.
After each convolutional layer, a batch normalization layer~\cite{ioffe2015batch} and no-linear activation function (ReLU) is applied. 
The max-pooling is used to reduce the spatial resolution with the kernel size of 2$\times$2 and the stride of 2.
We use the global average pooling (GAP)~\cite{lin2013network} to extract the global-context $f_g$ on the end of convolutional layers. 
We extract the local feature map $f_l$ after the max-pooling layer of the third block $f_3$ since it contains high-level abstract feature and has a good spatial resolution.
The local feature map $f_l$ interacts with the global context $f_g$ in the residual recurrent neural network (Residual RNN, as shown in Fig.~\ref{fig:framework}) block.
Finally, a fully-connected layer with softmax is used as a classifier for writer identification.

\subsection{Fragment-sequence learning}

\begin{figure}[!t]
    \centering
    \includegraphics[width=0.8\textwidth]{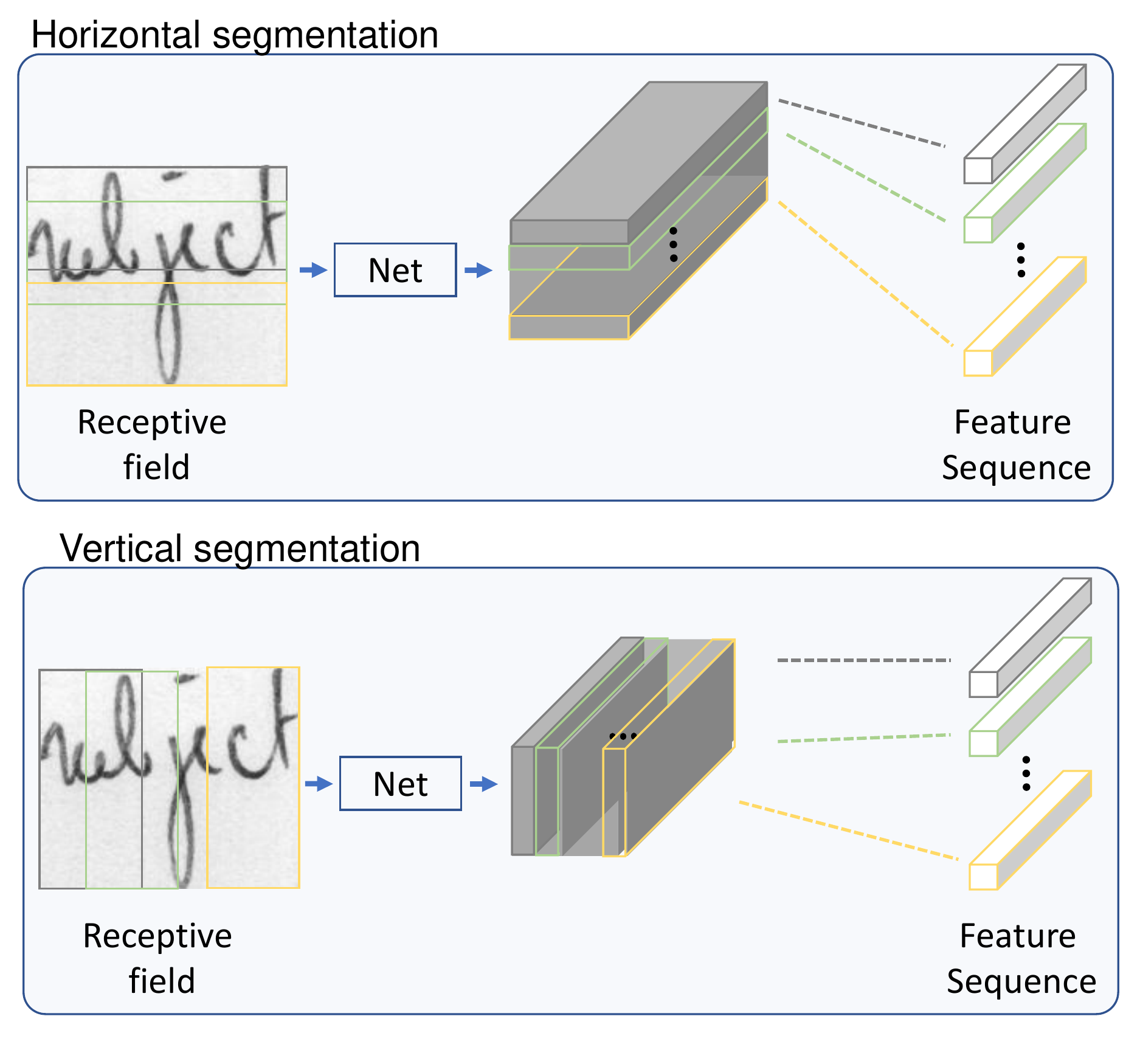}
    \caption{Two fragment segmentation methods on feature maps: horizontal segmentation (top figure) and vertical segmentation (bottom figure). Each feature vector in the feature sequence captures the writing style of the corresponding receptive field in the raw input handwritten image.}
    \label{fig:split}
\end{figure}

\begin{figure*}[!ht]
    \centering
    \includegraphics[width=\textwidth]{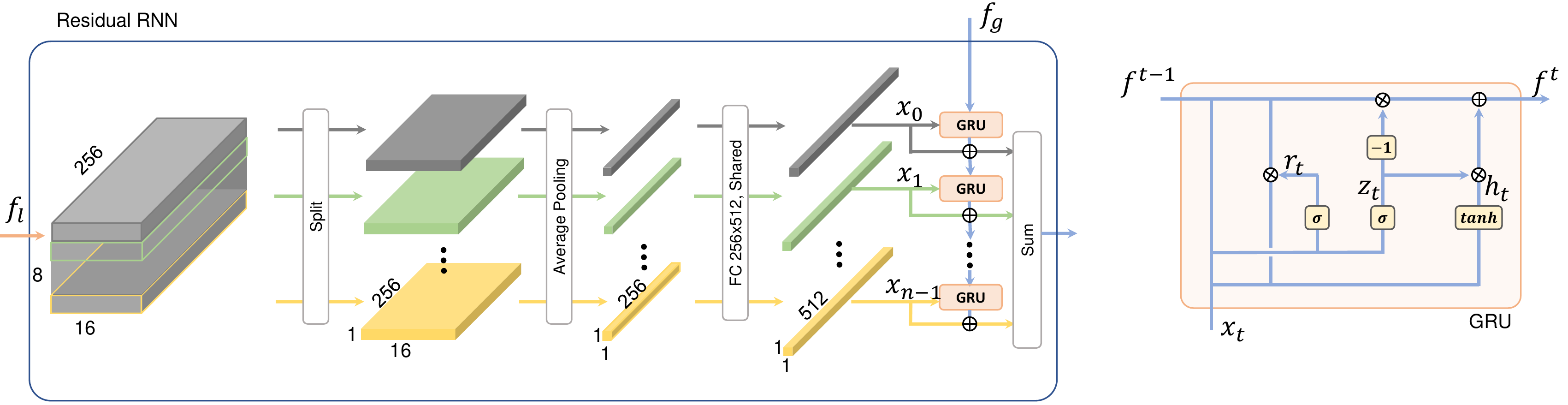}
    \caption{The left figure shows the structure of the Global-context Residual RNN block with the horizontal segmentation as an example and the right figure shows the standard GRU cell.}
    \label{fig:GRRNN}
\end{figure*}

Many deep learning methods use the part-based representation~\cite{li2017learning,sun2018beyond,bai2020deep} to learn the discriminative features for recognition based on the fact that 
the global feature pays more attention to the global information while the part-based features segmented from the deep learned feature maps care more about the local regions of the input image.
As shown in Fig.~\ref{fig:framework}, the global context $f_g$ contains global information of the handwriting style while ignores the local detailed information from fragments of the word (parts of the input image) which is very discriminative for writer identification.
The fragment-based representation is complementary to the global-context information $f_g$. 
Our previous work in~\cite{he2020fragnet} aggregates the local fragment features for writer identification and it provides a good performance. 
However, it simply combines the representation of fragments and the relationship between fragments has not been exploited. 
In this paper, we propose to use the recurrent neural network to naturally model the spatial dependency between each fragment segmented on the deep feature map $f_l$.
Instead of using zero values to initialize the hidden states of the RNN, we initialize the hidden states as the global-context $f_g$ in order to strengthen the discriminative ability of the deep features by combining the complementary global representation and the RNN-based local representation.

In order to exploit the relationship between local fragments, the feature map $f_l$, with the size of $8\times 16\times 256$ (height$\times$width$\times$depth), is decomposed into a sequence of fragments.
There are two different ways to segment the feature $f_l$ into a sequence of fragments (as shown in Fig~\ref{fig:split}): the horizontal segmentation which segments fragments along the height of the feature vector and the vertical segmentation which segments fragments along the width of the feature vector.
As shown in Fig.~\ref{fig:split}, each feature vector describes a rectangle region which contains a fragment of handwritten strokes in the raw input word image given by the corresponding receptive field.
The fragment segmented on the horizontal direction contains a single character, a part of character or the connection between characters, which carries the detailed handwriting style information.
The fragment segmented on the vertical direction represents the layout information of the handwritten word, which also contains useful information for writer identification~\cite{de2018layout}.
In this paper, the step of the horizontal segmentation is set to 1 and the step of the vertical segmentation is set to 2. 
Thus, the number of fragments is 8 for the two different segmentation ways and the size of each feature fragment is $1\times 16\times 256$ and $8\times 2\times 256$ for horizontal and vertical segmentation, respectively.

\subsection{Global-context Residual Recurrent Neural Network}
This section presents the details of the proposed residual RNN block which integrates the sequence of the fragments segmented from the local feature $f_l$ and the global-context information $f_g$.
Fig.~\ref{fig:GRRNN} shows the framework of the global-context residual RNN with the horizontal segmentation as an example.

To extract the feature vector of each fragment in the sequence, the global average pooling is used for each fragment to obtain the feature vector with the size of $1\times 1\times 256$. 
Then a fully-connected (FC) layer is used to increase the feature dimension from $256$ to $512$, which has the same feature dimension as the global-context feature $f_g$. Note that this FC layer is shared among all 8 fragments. 
Finally, a sequence of the feature vectors is obtained, denoted as $x_1,x_2,...,x_8$ with the dimension of 512.
The sequence of feature vectors is fed into the Gated Recurrent Units (GRU)~\cite{bengio1994learning} to model the spatial dependency between each fragment.
The GRU is an improved version of the simple recurrent neural network that can alleviate the vanishing and exploding gradient problems and it is defined (shown in the right figure of Fig.~\ref{fig:GRRNN}) as:
\begin{equation}
    \begin{split}
        z_t &= \sigma(W_zx_t+U_zf^{t-1}+b_z) \\
        r_t &= \sigma(W_rx_t+U_rf^{t-1}+b_r) \\
        h_t &= \phi(W_hx_t+U_h(r_t \circledast f^{t-1}) + b_h) \\
        f^t &= z_t \circledast f^{t-1} + (1-z_t) \circledast h_t
    \end{split}
\end{equation}
where $x_t$ is the fragment feature vector and $f^t$ is the global context feature at the time step $t$ ($f^0=\vec{0}$ for the vanilla GRU and $f^0$=$f_g$ for the proposed method as shown in Fig.~\ref{fig:framework}).
$z_t$ and $r_t$ are the update gate and reset gate with the sigmoid $\sigma(x)$ as the activation function.
$\phi(x)$ is the hyperbolic tangent function and $W_*, U_*, b_*$ are the parameters learned during training.
The GRU can be denoted as $f^t=\text{GRU}(x_t,f^{t-1})$.

Note that unlike the vanilla GRU where the hidden units are initialized by zero values, the proposed method considers the global-context feature $f_g$ as the initialization of the hidden units.
Thus, the proposed method is the global-context RNN and the global context feature $f_g$ is updated recurrently with the local fragment features $x_t, t\in\{1,2,...,8\}$.
The irrelevance information in $f^t$ is suppressed and the important information is enhanced and supplemented from the local feature $x_t$ by the two update and resent gates in GRU.

In order to fully exploit the handwriting style information in the local fragment features, we propose the global-context residual GRU, inspired by the residual network~\cite{he2016deep}, as following:
\begin{equation}
\label{eq:res}
    f^t = \text{GRU}(x_t, f^{t-1})+x_t \ \ \ \ t\in\{1,2,\cdots,8\}, f^0 = f_g
\end{equation}
where the feature vector $f^t$ is initialized by the global-context $f_g$, updated and enhanced by RNN with the local feature $x_t$ at the time step $t$.
Thanks to the context modeling of GRU, each resulting feature vector $f^t$ contains the global context and local discriminative information and thus describes the handwriting style better than its associated fragment $x_t$. 
Finally, the final feature vector $f$ of the input image is computed by:
\begin{equation}
    f = \sum_{t=1}^{8} f^t
\end{equation}
followed by a classifier with the softmax layer containing $N$ output neurons, corresponding to the $N$ writer identities.

\subsection{Label smooth regularization}
Label smooth regularization was proposed in~\cite{szegedy2016rethinking} and has been used for deep feature learning~\cite{chen2019semi} to mitigate the overfitting problem and to make the model more adaptable.
A new ground-truth label is defined as a mixture of the original one-hot ground-truth distribution $p$ and a fixed distribution $u$:
\begin{equation}
    \hat{p} = (1-\epsilon) * p + \epsilon u
\end{equation}
where $\epsilon$ is the smoothing parameter. 
Following~\cite{szegedy2016rethinking,chen2019semi}, we use the uniform distribution $u=1/N$ where $N$ is the number of writers and $\epsilon = 0.1$.
Finally, the training loss is defined as:
\begin{equation}
    loss = -(1-\epsilon) * \text{log}(p_y) - \frac{\epsilon}{N}\sum_{n=1}^{N}\text{log}(p_n)
\end{equation}
where $y$ is the ground-truth of the writer identity.
The loss not only pays attention to the ground-truth $y$ with a high weight $1-\epsilon+\epsilon/N>0.9$ but also takes other classes with a low weight $\epsilon/N$ into account.

\section{Experimental results}
\label{sec:exps}

\subsection{Data set}

\begin{table}[!t]
	\renewcommand{\arraystretch}{1.3}
	\caption{The number of training and testing word images on each data set.}
	\label{tab:dataset}
	\centering
	\begin{tabular}{l|c|c|c}
		\toprule[1pt]
		Data set & \#Writers  & \#Training & \#Testing\\
		\midrule
		IAM~\cite{marti2002iam} 				& 657 	& 56,432 	& 25,827 \\
		CVL~\cite{kleber2013cvl}				& 310 	& 62,406	& 34,564\\
		Firemaker~\cite{schomaker2000forensic}	& 250 	& 25,256	& 11,595	\\
		CERUG-EN~\cite{he2015junction}			& 105 	& 5,702		& 5,127 	\\
		\bottomrule[1pt]
	\end{tabular}
\end{table}

We use the IAM~\cite{marti2002iam}, 
CVL~\cite{kleber2013cvl}, Firemaker~\cite{schomaker2000forensic} and 
CERUG-EN~\cite{he2015junction} datasets to evaluate the performance of writer identification based on word images.
All datasets, with the splitting of training and testing sets, are public available on the website~\footnote{\href{https://www.ai.rug.nl/~sheng/writeridataset.html}{\url{https://www.ai.rug.nl/~sheng/writeridataset.html }}}.
IAM~\cite{marti2002iam} contains 1,452 document images, segmenting into 82,259 word images from 657 writers.
The distribution of writers on IAM is unbalanced.
A half of writers contribute several pages and the rest of writers have only one page or even a few sentences.
The CVL~\cite{kleber2013cvl} dataset contains 310 writers with 96,970 word images in German and English.
There are 250 and 105 writers on the Firemaker and CERUG-EN datasets, respectively.
Table~\ref{tab:dataset} shows the number of training and testing samples on the four datasets.
Word images from one page appear only in the training or the testing set, making it possible to evaluate performance of writer identification based on line-level or page-level images.
More detailed information of these datasets can be found in~\cite{he2020fragnet}.

\subsection{Implementation details}

The proposed GR-RNN model is built on the PyTorch framework.
Neural networks are trained with the Adam optimizer~\cite{kingma2014adam}. 
The weight decay is set to 0.0001 and the mini-batch size is set to 16.
The initial learning rate is set to 0.0001 and a decay schedule (reduce to half) at every 10 epochs is applied. 
The model is trained with 50 epochs. We resize all word images to a fixed size $(64,128)$ for training the proposed neural network, similar to our previous works~\cite{he2019deep,he2020fragnet}. 
Word images are resized by keeping the aspect ratio without distortions and
padding is used when it is necessary.
We use the simple translation augmentation method to avoid positional bias in the data during training.

\begin{table*}[!t]
    \caption{Different methods used to compute the feature vector $f$, followed by a classifier with Softmax. $f_g$ is the global context and $x_t$ is the sequence of the fragments segmented on feature map $f_l$. The number of parameters and FLOPs is computed based on the IAM dataset (with 657 writers). }
    \label{tab:models}
    \centering
    \resizebox{\textwidth}{!}{
    \begin{tabular}{l|l|l|c|c}
    \toprule
         Model & Feature computation & Description & Parameters (MB) & FLOPs (G)\\
    \midrule
        Global-context (Baseline) & $f=f_g$  & The typical CNN model with the GAP & 5.15 & 1.67\\
        Fragment-only (F) & $f=\sum x_t$ & Assemble local fragments & 6.73 & 1.21 \\
        Fragment-RNN (FR) & $f=\sum f^t, f^t=\text{GRU}(x_t,f^{t-1}), f^0=\vec{0}$ & Vanilla RNN & 6.73 & 1.24 \\
        Fragment-Residual-RNN (FRR) & $f=\sum f^t, f^t=\text{GRU}(x_t,f^{t-1})+x_t, f^0=\vec{0}$ & Residual RNN & 6.73 & 1.24 \\
        Fragment-Global-context-RNN (FGR) & $f=\sum f^t, f^t=\text{GRU}(x_t,f^{t-1}), f^0=f_g$ & Vanilla RNN with global-context & 6.73 & 1.69\\
        Fragment-Globabl-context-Residual-RNN (FGRR) & $f=\sum f^t, f^t=\text{GRU}(x_t,f^{t-1})+x_t, f^0=f_g$ & Residual RNN with global-context & 6.73 & 1.69 \\
    \bottomrule
    \end{tabular}}
    
\end{table*}

\subsection{Compared models}

We compare different models described in Table~\ref{tab:models} and the main difference is how to compute the feature vector $f$ which is fed into the classifier with softmax layer for writer identification:
(1) Global-context (Baseline): a classifier is built with the feature vector $f_g$ of the global-context computed by the global average pooling. 
This is baseline which has the similar structure with the traditional neural network without the Residual RNN block shown in Fig.~\ref{fig:framework}.
(2) Fragment-only (F): a classifier is built with the feature vector of the sum of the fragments segmented from the feature map $f_l$.
This is similar to use the global average pooling on the feature map $f_l$.
(3) Fragment-RNN (FR): applying the vanilla RNN on the sequence of the fragments with zero-initialized hidden units.
(4) Fragment-Residual-RNN (FRR): using residual RNN defined in Eq.~\ref{eq:res} on the sequence of the fragments with zero-initialized hidden units.
(5) Fragment-Global-context-RNN (FGR): using the RNN on the sequence of the fragments with the global context as the initialization of the hidden units.
(6) Fragment-Globabl-context-Residual-RNN (FGRR): the proposed global-context residual RNN model, applying the residual RNN with the global context as the initialization of the hidden units.
We also compare these different networks with the horizontal and vertical fragment segmentation methods described in Fig.~\ref{fig:split}.
The number of parameters and the corresponding floating point operations (FLOPs) of each model (on the IAM dataset with 657 writers) are also reported in Table~\ref{tab:models}.

\subsection{Writer identification on different images}

\begin{figure}[!t]
    \centering
    \includegraphics[width=0.8\textwidth]{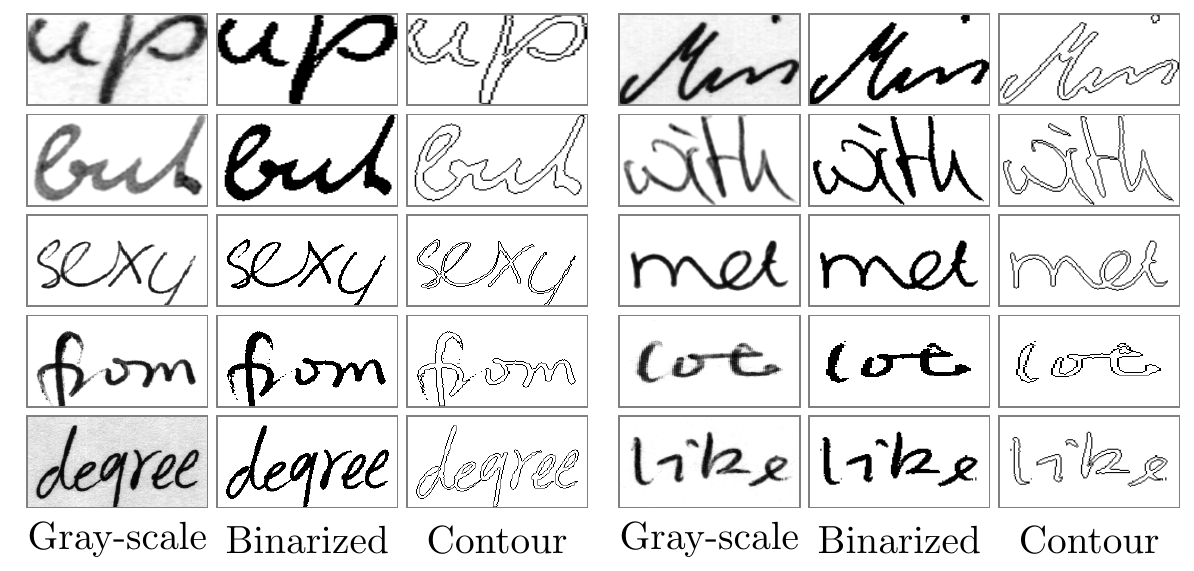}
    \caption{The gray-scale images and their corresponding binarized and contour images.}
    \label{fig:imgbincon}
\end{figure}

One widely accepted intuition for writer identification is that the handwriting style is encapsulated on the shape of the handwritten texts, especially on the contours of the ink traces~\cite{bulacu2007text,siddiqi2010text,brink2012writer,he2017writer}. 
Thus, the handcrafted features extracted on the contours of handwritten texts have been successfully used for writer identification based on page images.
However, it has shown in~\cite{geirhos2018imagenet} that convolutional neural networks are strongly biased towards recognizing objects by textures rather than shapes.
In order to evaluate whether convolutional neural networks also pay attention to shapes or contours of the handwritten texts for writer identification, we use different types of input images, such as gray-scale, binarized and contour images to train the neural network.
Fig.~\ref{fig:imgbincon} shows several examples of these images.
Given the gray-scale image, we use the global Otsu~\cite{otsu1979threshold} threshold to get the binarized image~\cite{christlein2017writer}. 
Then handwritten contours are extracted based on the binarized images using the tracking method proposed in~\cite{brink2012writer}.
All images are normalized into floating point values in the range of [0,1] and
neural networks are trained with the same configuration for fair comparison.

\begin{figure}[!t]
    \centering
    \includegraphics[width=\textwidth]{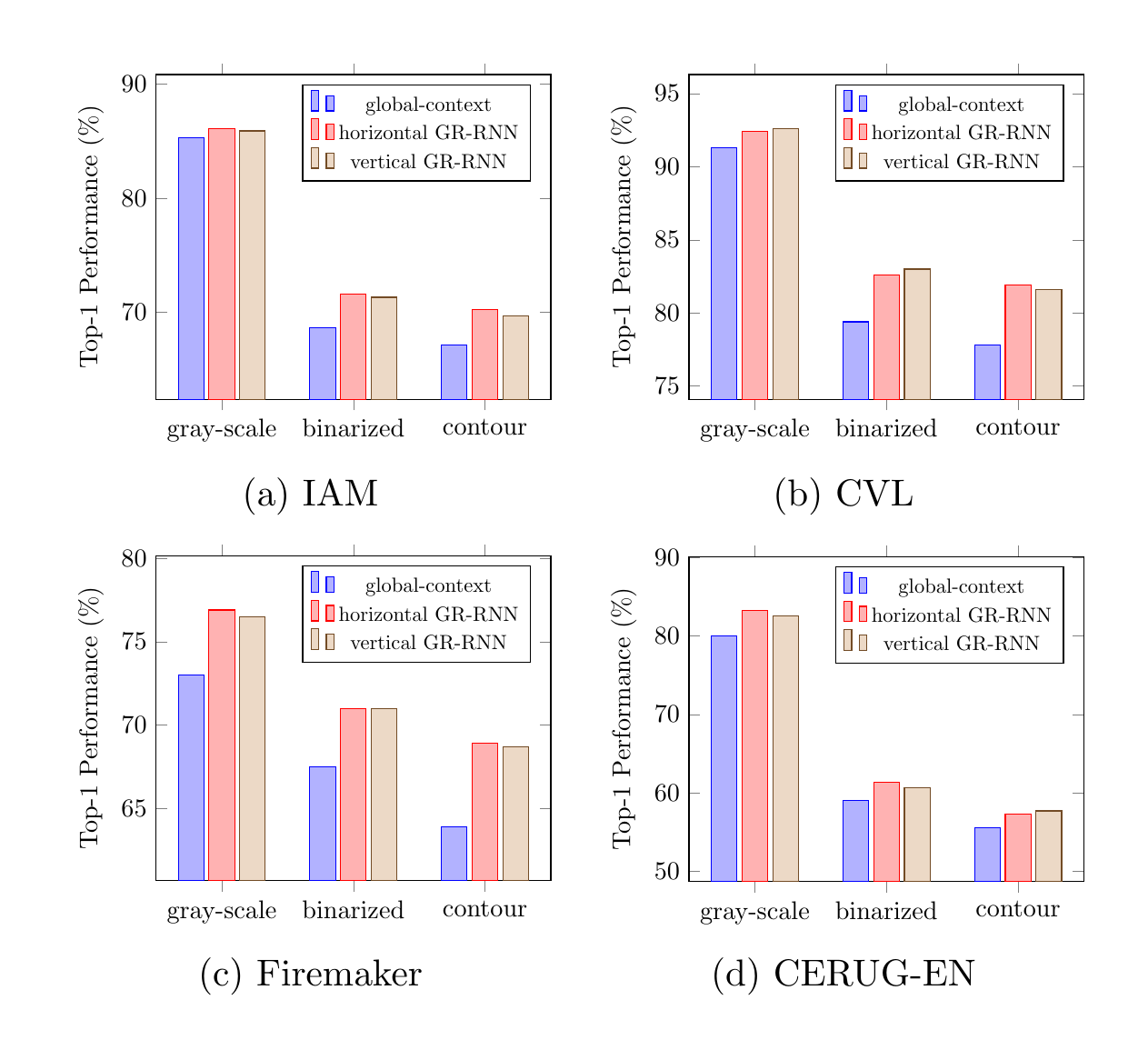}
    \caption{The performance of different types of images for writer identification on four datasets.}
    \label{fig:imgmodes}
\end{figure}

Fig.~\ref{fig:imgmodes} shows the performance of writer identification using different neural networks on different types of images on the four datasets.
There are marked performance differences among neural networks trained with different images.
CNNs trained on the gray-scale images provide the best performance since the gray-scale images contain rich texture information on the ink traces caused by the writing instruments and the writer's pressures during writing.
The performance on binarized images is slightly better than performance on contour images.
The reason might be that binarized images contain more ink pixels than contour images, thus more valid information is kept in the neural network after several max-pooling layers.

The results indicate that the CNN tends to recognize writers according to the local textures which contain the subtle information of writer identity.
In addition, with the same image, the proposed GR-RNN method provides better results than global-context neural networks (the baseline) on four datasets.
Our results suggest that the texture information plays an important role, which provides a new insight for forensic examiners to consider the texture information of the ink traces for writer identification in future.

In the following sections, we conduct experiments of writer identification using the gray-scale images.

\subsection{Performance of writer identification based on the gray-scale word images}
\begin{figure}[!t]
    \centering
    \includegraphics[width=\textwidth]{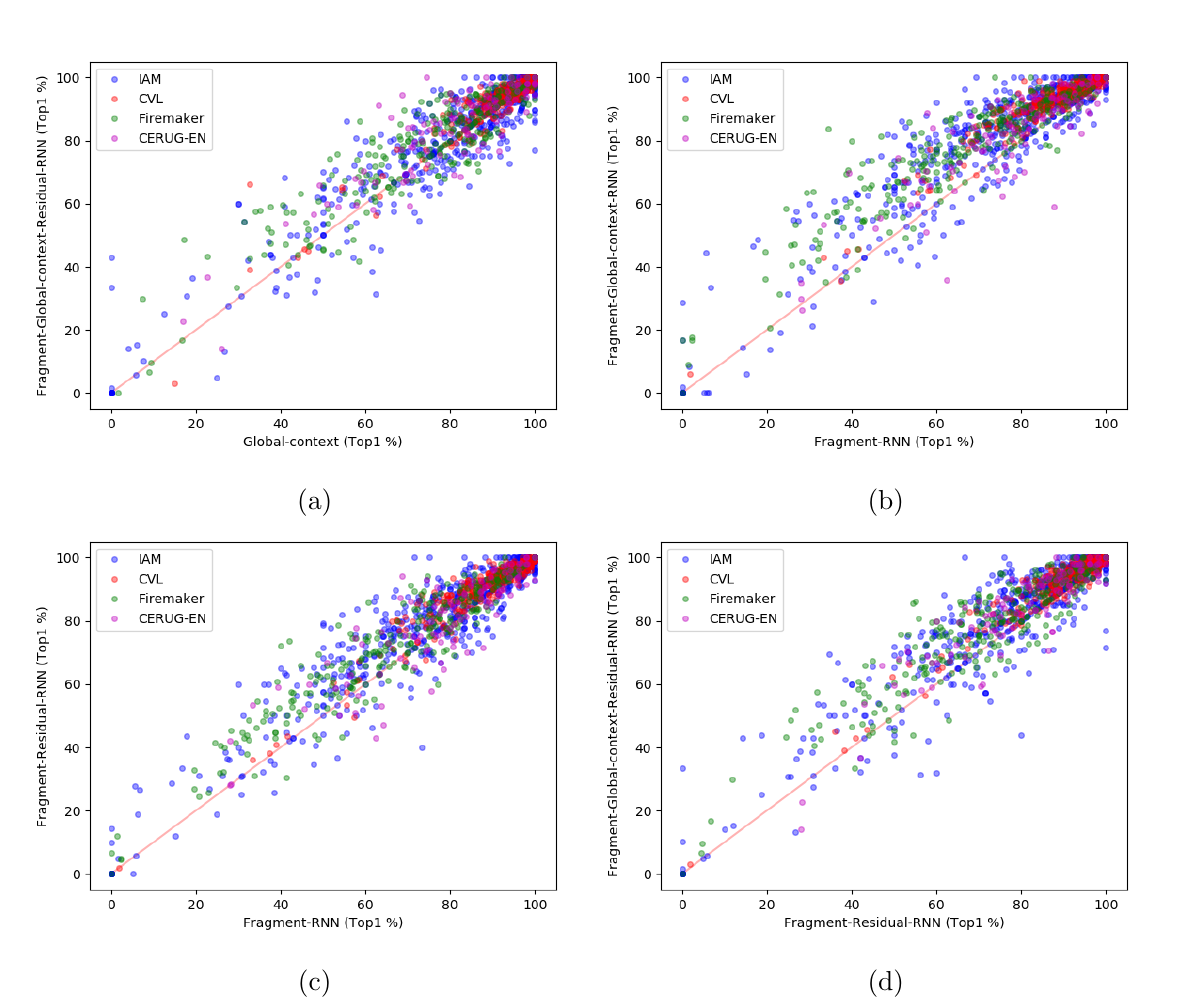}
    \caption{Scatter Plots of writer-conditional Top-1 accuracy in the four datasets. (a) FGRR against Global-context. (b) FGR against FR. (c) FRR against FR. (d) FGRR against FRR. Note we only plot the results of horizontal segmentation and a similar trend is found in the results of different networks with the vertical segmentation method.}
    \label{fig:scatter}
\end{figure}

In this section, we train different neural networks listed on Table~\ref{tab:models} to evaluate the performance for writer identification with different fragment segmentation methods.
Table~\ref{tab:wordres} shows the performance of the different models based on the fragment sequence of vertical and horizontal segmentation for writer identification on word images.
Fig.~\ref{fig:scatter} shows the scatter plots of top-1 accuracy within each writer between different models. 
We can see that using the vanilla RNN on the sequence of fragments (FR) segmented on the low-level feature space provides the worst results.
However, combining the residual information (FRR) can improve the performance for most writers (Fig.~\ref{fig:scatter}(c)).
The performance is further boosted by integrating the global-context information (FGR and FGRR) (Fig.~\ref{fig:scatter}(b) and (d)).
In general, models built on the sequence of fragments with global-context (FGR, FGRR) give better results than models built on the sequence of fragments without the global-context (F, FR, FRR).
In most cases, the performance of our proposed GR-RNN (FGRR) is better than baseline (Fig.~\ref{fig:scatter} (a)) and other neural networks.
We can also find that there is no significant difference between models using horizontal or vertical fragment segmentation methods.

\begin{table}[!t]
	\renewcommand{\arraystretch}{1.3}
	\caption{The writer identification performance of different neural networks using the horizontal and vertical segmentation based on word images.}
	\label{tab:wordres}
	\centering
	\resizebox{\textwidth}{!}{
	\begin{tabular}{cl|cc|cc|cc|cc}
		\toprule[1pt]
		\multicolumn{2}{c}{\multirow{2}{*}{Dataset}} & \multicolumn{2}{|c|}{IAM} & \multicolumn{2}{c|}{CVL} & \multicolumn{2}{c|}{Firemaker} &
		\multicolumn{2}{c}{CERUG-EN}\\
		\cmidrule{3-10}
		& & Top1 & Top5 & Top1 & Top5 & Top1 & Top5 & Top1 & Top5\\
		\midrule
		
		\multicolumn{2}{c|}{Baseline}   & 85.3 & 95.2 & 91.3 & 97.6 & 73.0 & 89.5 & 80.0 & 96.4\\
		\midrule
		\multirow{5}{*}{\rotatebox{90}{Vertical}}
		& F & 82.2 & 93.7 & 89.4  & 97.0 & 70.3 & 88.2 & 78.9 & 95.3\\
		& FR & 80.1 & 92.8 & 87.8 & 96.4 & 65.7 & 85.8 & 75.0 & 94.6\\
		& FRR & 83.0 & 93.9 & 90.3 & 97.2 & 72.6 & 89.4 & 79.2 & 95.3  \\
		& FGR & 86.3 & 95.1 & 92.4 & 97.8 & 75.7 & 90.8 & 80.5 & 95.1 \\
		& FGRR & 85.9 & 95.2 & 92.6 & 97.9 & 76.5 & 91.1 & 82.6 & 95.8 \\ 
		\midrule
		\multirow{5}{*}{\rotatebox{90}{Horizontal}}
		& F & 82.1 & 93.6 & 89.1 & 96.8 & 70.7 & 88.7 & 80.8 & 96.2 \\
		& FR & 80.1 & 92.9 & 87.0 & 96.2 & 65.2 & 85.2 & 77.7 & 95.6 \\
		& FRR & 82.9 & 94.0 & 89.8 & 97.0 & 70.7 & 88.5 & 79.4 & 95.3 \\
		& FGR & 85.3 & 94.8 & 92.1 & 97.7 & 75.9 & 90.5 & 81.9 & 95.9 \\
		& FGRR & 86.1 & 95.0 & 92.4 & 97.8 & 76.9 & 91.0 & 83.2 & 96.2 \\ 
		\bottomrule[1pt]
	\end{tabular}}
\end{table}

\subsection{Performance comparison with previous studies}

Table~\ref{tab:comparison} shows the comparison with the ResNet~\cite{he2016deep} and the FragNet-$64$ method~\cite{he2020fragnet}.
It can be observed from the table that the proposed GR-RNN model has less parameters than FragNet-$64$ which has two pathways: global feature pyramid and fragment pathway.
The GR-RNN has only one pathway and the sequence of fragments is segmented on the feature map.
From the table we can see that our proposed methods improve the performance in terms of the Top-1 recognition rate on the four datasets.

\begin{table}[!t]
	\renewcommand{\arraystretch}{1.3}
	\caption{Comparison with other methods for writer identification based on word images.}
	\label{tab:comparison}
	\centering
	\resizebox{\textwidth}{!}{
	\begin{tabular}{l|c|cc|cc|cc|cc}
		\toprule[1pt]
		\multirow{2}{*}{Dataset} & \multirow{2}{*}{\#params} & \multicolumn{2}{c|}{IAM} & \multicolumn{2}{c|}{CVL} & \multicolumn{2}{c|}{Firemaker} &
		\multicolumn{2}{c}{CERUG-EN}\\
		\cmidrule{3-10}
		& & Top1 & Top5 & Top1 & Top5 & Top1 & Top5 & Top1 & Top5\\
		\midrule
		ResNet18~\cite{he2016deep,he2020fragnet} & 11.3M & 83.2 & 94.3 & 88.5 & 96.7 & 63.9 & 86.4 & 70.6 & 94.0 \\
		FragNet-64~\cite{he2020fragnet} & 18.2M & 85.1 & 95.0 & 90.2 & 97.5 & 69.0 & 88.5 & 77.5 & 95.6 \\
		Vertical GR-RNN (FGRR)   & 6.9M & 85.9 & 95.2 & 92.6 & 97.9 & 76.5 & 91.1 & 82.6 & 95.8\\
		Horizontal GR-RNN (FGRR) & 6.9M & 86.1 & 95.0 & 92.4 & 97.8 & 76.9 & 91.0 & 83.2 & 96.2\\
		\bottomrule[1pt]
	\end{tabular}}
	
\end{table}

\subsection{Performance comparison with different features}

We also evaluate the features $f$ learned by the proposed GR-RNN method for writer identification, comparing with other handcrafted features.
The feature vector $f$ is normalized by $\hat{f}=f/||f||_2$.
For each writer, the model is built as the average feature of the all word images from the same writer on the training set. 
The nearest neighbor method with the Euclidean distance is used for writer identification given the query word images on the testing set.
Table~\ref{tab:compareOthers} shows the performance of different methods.
We do not compare the traditional grapheme-based features since the information in a single word image is not sufficient to build a stable feature vector for capturing the writing style based on the bag-of-word model.
This can be found in~\cite{brink2008much} that at least 150 characters are required to achieve the very satisfactory results.
Generally, the deep learned features provide much better results than handcrafted features for writer identification based on word images.
Our proposed GR-RNN provides the best results on the IAM, CVL and Firemaker datasets.
However, it gives worse results than FragNet-$64$ on CERUG-EN which has less training samples than other three datasets (as shown in Table~\ref{tab:dataset}).

\begin{table}[!t]
	\centering 
	\renewcommand{\arraystretch}{1.3}
	\caption{The comparison of writer identification performance based on \textbf{word images} using different features.}
	\label{tab:compareOthers}
	\resizebox{\textwidth}{!}{
	\begin{tabular}{l|cc|cc|cc|cc}
		\toprule[1pt]
		\multirow{2}{*}{Method} & \multicolumn{2}{c|}{IAM} & \multicolumn{2}{c|}{CVL} & \multicolumn{2}{c|}{Firemaker} & \multicolumn{2}{c}{CERUG-EN} \\
		\cline{2-9}
													& Top1 & Top5 & Top1 & Top5 & Top1 & Top5 & Top1 & Top5 \\
		 \midrule
		Hinge~\cite{bulacu2007text} 				& 13.8 & 28.3 & 13.6 & 29.7 & 19.6 & 40.0 & 14.4 & 32.8\\
		Quill~\cite{brink2012writer}				& 23.8 & 44.0 & 23.8 & 46.7 & 21.7 & 43.7 & 24.5 & 51.9 \\
		CoHinge~\cite{he2017beyond}					& 19.4 & 34.1 & 18.2 & 34.2 & 27.4 & 48.2 & 17.7 & 34.0\\
		QuadHinge~\cite{he2017beyond} 				& 20.9 & 37.4 & 17.8 & 35.5 & 26.5 & 47.4 & 17.0 & 36.0\\
		COLD~\cite{he2017writer}					& 12.3 & 28.3 & 12.4 & 29.0 & 22.7 & 45.1 & 17.3 & 42.2 \\
		Chain Code Pairs~\cite{siddiqi2010text} 	& 12.4 & 27.1 & 13.5 & 30.3 & 17.5 & 36.8 & 14.5 & 33.0\\
		Chain Code Triplets~\cite{siddiqi2010text}	& 16.9 & 33.0 & 17.2 & 35.4 & 22.9 & 43.8 & 17.8 & 38.0\\
		FragNet-$64$~\cite{he2020fragnet} & 72.2 & 88.0 & 79.2 & 93.3 & 57.5 & 80.8 & 75.9 & 94.7\\
		\midrule
		Vertical GR-RNN (FGRR) &  83.3 & 94.0 & 83.5 & 94.6 & 60.8 & 83.2 & 70.2 & 91.6\\
		Horizontal GR-RNN (FGRR) & 82.4 & 93.8 & 82.9 & 94.6 & 62.2 & 83.9 & 68.9 & 90.9  \\
		\bottomrule
	\end{tabular}}
\end{table}

\subsection{Performance on line and page images}

Table~\ref{tab:lineres} and Table~\ref{tab:pageres} show the performance of different models for writer identification based on line and page images, respectively.
The writer evidence is computed as the average response of all word images on each line image and page image.
From these tables, the similar conclusions can be found among the performance of different neural works on the word, line and page images.
The performance of writer identification based on page images is higher than performance based on word and line images since each page image contains several line images and each line contains several word images.
In addition, there is no significant difference among the performance of different neural networks on page level (as shown in Table~\ref{tab:pageres}).
Thus, given a large amount of training data, the basic neural network with several convolutional layers can provide satisfactory performance.

\begin{table}[!t]
	\renewcommand{\arraystretch}{1.3}
	\caption{he writer identification performance of different neural networks using the horizontal and vertical segmentation based on \textbf{line images}.}
	\label{tab:lineres}
	\centering
	\resizebox{\textwidth}{!}{
	\begin{tabular}{cl|cc|cc|cc|cc}
		\toprule[1pt]
		\multicolumn{2}{c}{\multirow{2}{*}{Dataset}} & \multicolumn{2}{|c|}{IAM} & \multicolumn{2}{c|}{CVL} & \multicolumn{2}{c|}{Firemaker} &
		\multicolumn{2}{c}{CERUG-EN}\\
		\cmidrule{3-10}
		& & Top1 & Top5 & Top1 & Top5 & Top1 & Top5 & Top1 & Top5\\
		\midrule
		
		\multicolumn{2}{c|}{Baseline}   & 94.5 & 97.4 & 99.4 & 99.9 & 94.9 & 98.7 & 94.1 & 99.2 \\
		\midrule
		\multirow{5}{*}{\rotatebox{90}{Vertical}}
		& F & 94.3 & 97.4 & 99.4 & 99.9 & 94.2 & 98.2 & 94.6 & 98.8 \\
		& FR & 93.4 & 97.2 & 99.4 & 99.8 & 93.7 & 97.9 & 93.5 & 98.5 \\
		& FRR & 94.5 & 97.5 & 99.7 & 99.9 & 96.2 & 98.8 & 94.4 & 98.5  \\
		& FGR & 95.3 & 97.5  & 99.5 & 99.7 & 95.9 & 98.8 & 95.8 & 99.4 \\
		& FGRR &  95.2 & 97.4 & 99.5 & 99.9 & 96.7 & 98.8 & 96.6 & 99.6 \\ 
		\midrule
		\multirow{5}{*}{\rotatebox{90}{Horizontal}}
		& F &  94.3 & 97.4 & 99.4 & 99.7 & 95.2 & 98.9 & 95.0 & 99.2 \\
		& FR & 93.5 & 97.2 & 99.4 & 99.7 & 92.7 & 97.3 & 94.8 & 99.0 \\
		& FRR & 94.7 & 97.3 & 99.5 & 99.7 & 94.9 & 97.9 & 95.8 & 99.4\\
		& FGR &  94.8 & 97.4 & 99.6 & 99.9 & 95.9 & 98.6 & 95.2 & 99.6\\
		& FGRR &  95.1 & 97.5 & 99.6 & 99.8 & 96.3 & 98.7 & 95.8 & 99.6\\ 
		\bottomrule[1pt]
	\end{tabular}}
\end{table}

\begin{table}[!t]
	\renewcommand{\arraystretch}{1.3}
	\caption{he writer identification performance of different neural networks using the horizontal and vertical segmentation based on \textbf{page images}.}
	\label{tab:pageres}
	\centering
	\resizebox{\textwidth}{!}{
	\begin{tabular}{cl|cc|cc|cc|cc}
		\toprule[1pt]
		\multicolumn{2}{c}{\multirow{2}{*}{Dataset}} & \multicolumn{2}{|c|}{IAM} & \multicolumn{2}{c|}{CVL} & \multicolumn{2}{c|}{Firemaker} &
		\multicolumn{2}{c}{CERUG-EN}\\
		\cmidrule{3-10}
		& & Top1 & Top5 & Top1 & Top5 & Top1 & Top5 & Top1 & Top5\\
		\midrule
		
		\multicolumn{2}{c|}{Baseline}   & 95.9 & 98.2 & 99.3 & 99.6 & 98.4 & 100.0 & 98.1 & 100.0\\
		\midrule
		\multirow{5}{*}{\rotatebox{90}{Vertical}}
		& F & 95.9 & 98.3 & 99.1 & 99.4 & 98.0 & 100.0 & 98.1 & 100.0 \\
		& FR & 95.7 & 98.2 & 99.4 & 99.4 & 98.8 & 99.6 & 98.1 & 100.0 \\
		& FRR &  96.4 & 98.3 &  99.4 & 99.7 & 99.2 & 99.6 & 97.1 & 100.0 \\
		& FGR &  96.6 & 98.3 & 99.3 & 99.4 & 98.4 & 100.0 & 99.1 & 100.0 \\
		& FGRR &  96.4 & 98.3 & 99.3 & 99.4 & 98.8 & 100.0 & 99.1 & 100.0\\ 
		\midrule
		\multirow{5}{*}{\rotatebox{90}{Horizontal}}
		& F &  95.0 & 98.2 & 99.4 & 99.4 & 98.8 & 100.0 & 99.1 & 100.0\\
		& FR & 95.6 & 98.2 & 99.1 & 99.4 & 97.2 & 99.2 & 98.1 & 100.0\\
		& FRR & 96.5 & 98.2 & 99.1 & 99.4 & 98.4 & 98.8 & 99.1 & 100.0 \\
		& FGR &  96.2 & 98.3 & 99.4 & 99.6 & 98.8 & 99.6 & 99.1 & 100.0 \\
		& FGRR &  96.4 & 98.3 & 99.3 & 99.4 & 98.8 & 100.0 & 98.1 & 100.0\\ 
		\bottomrule[1pt]
	\end{tabular}}
\end{table}

\subsection{Performance comparison with state-of-the-art methods}

Table~\ref{tab:SOA} presents the comparison with state-of-the-art methods for writer identification based on page images.
Although different methods in this table may use different experimental protocols, it is still interesting to draw some conclusions.
For IAM dataset, the traditional handcrafted feature~\cite{wu2014offline} provides the best results than the deep learning methods. The possible reason might be that the number of samples for each writer is unbalanced. 
Some writers contributed several pages and some only wrote one page with several sentences. 
Our proposed method provides better results than deep learning methods~\cite{nguyen2019text,he2020fragnet} on the IAM dataset.
The CVL and CERUG-EN datasets are the easy datasets for writer identification and our proposed method provides a comparative results with other studies.
The proposed method provides the best performance on the Firemaker dataset.

\begin{table}[!t]
    \centering
    \caption{Comparison of state-of-the-art methods on the four datasets.}
    \label{tab:SOA}
    \resizebox{\textwidth}{!}{
    \begin{tabular}{p{0.5cm}lclc}
    \toprule[1pt]
        Dataset &  Reference & \#Writer & Feature & Top-1(\%)  \\
         \midrule
        \multirow{9}{*}{\rotatebox{90}{IAM}} & Siddiqi and Vincent~\cite{siddiqi2010text} & 650 & Contour and codebook features & 91.0 \\
        &He and Schomaker~\cite{he2017beyond} & 650 & Best results among 17 handcraft features & 93.2 \\
        &Khalifa et al.~\cite{khalifa2015off} & 650 & Graphemes with codebook & 92.0 \\
        &Hadjadji and Chibani~\cite{hadjadji2018two} & 657 & LPQ, RL and oBIF with OC-K-Means & 94.5 \\
        &Wu et al.~\cite{wu2014offline} & 657 & Scale invariant feature transform & \textbf{98.5} \\
        & Lai et al.~\cite{lai2020encoding} & 650 & Pathlet + bVLAD & \textbf{98.5} \\
        &Khan et al.~\cite{khan2018dissimilarity} &  650 & SIFT+RootSIFT & 97.8 \\
        &Nguyen et. al~\cite{nguyen2019text} & 650 & CNN with sub-images of size $64\times 64$ & 93.1 \\
        &FragNet-$64$~\cite{he2020fragnet} & 657 & CNN with word images and fragments & 96.3 \\
        &Proposed & 657 & Global-context residual recurrent neural network & 96.4\\
        \midrule
        \multirow{7}{*}{\rotatebox{90}{CVL}} & Fiel and Sablatnig~\cite{fiel2013writer} & 309 & SIFT with GMM & 97.8 \\
        &Tang and Wu~\cite{tang2016text} & 310 & CNN with joint Bayesian & \textbf{99.7} \\
        & Lai et al.~\cite{lai2020encoding} & 310 & Pathlet+bVLAD & 99.76 \\
        &Christlein et al.~\cite{christlein2017writer} & 310 & SIFT with GMM and Examplar-SVMs & 99.2 \\
        &Khan et al.~\cite{khan2018dissimilarity} &  310 & SIFT+RootSIFT & 99.0 \\
        & Chen et al.~\cite{chen2019semi} & 310 & Semi-supervised with ResNet-50 trained with IAM & 99.2 \\
        &FragNet-$64$~\cite{he2020fragnet} & 310 & CNN with word images and fragments & 99.1 \\
        &Proposed & 310 & Global-context residual recurrent neural network & 99.3 \\
        \midrule
          \multirow{6}{*}{\rotatebox{90}{Firemaker}} &He and Schomaker~\cite{he2017beyond} & 250 & Best results among 17 handcraft features & 92.2 \\
         &Wu et al.~\cite{wu2014offline} & 250 & Scale invariant feature transform & 92.4 \\
        &Khan et al.~\cite{khan2018dissimilarity} &  250 & SIFT+RootSIFT & 98.0 \\
        &Nguyen et. al~\cite{nguyen2019text} & 250 & CNN with sub-images of size $64\times 64$ &93.6 \\
         & Lai et al.~\cite{lai2020encoding} & 250 & Pathlet+bVLAD & 97.98 \\
        &FragNet-$64$~\cite{he2020fragnet} & 250 & CNN with word images and fragments & 97.6 \\
        & Proposed &  250 & Global-context residual recurrent neural network & \textbf{98.8} \\
         \midrule
        \multirow{3}{*}{\rotatebox{90}{CERUG}}& He and Schomaker~\cite{he2017beyond} & 105 & Best results among 17 handcraft features & 97.1 \\
        & FragNet-$64$~\cite{he2020fragnet} & 105 &CNN with word images and fragments & \textbf{100.0} \\
         & Proposed & 105 & Global-context residual recurrent neural network & 99.1 \\
         \bottomrule[1pt]
    \end{tabular}}
\end{table}

\section{Conclusion}
\label{sec:cons}

In this paper, we have introduced a novel global-context residual recurrent neural network framework for writer identification, which jointly employs the global context and the local fragment-part information.
The advantage of the proposed method is that it can capture the global-context information extracted by the convolutional neural network and local fine-grained information extracted by the recurrent neural network.
Thus, it can extract the detailed writing style information from one single word image and it achieves state-of-the-art performance on four public datasets.
The limitation of the proposed method is that it needs the word or sub-word image segmentation, which requires extra prepossessing steps for applying it on other documents (such as historical document fragments~\cite{seuret2020icfhr}). 
In future work, the proposed method can be extended to capture the handwriting style information on any handwritten document without any segmentation by using a sliding window strategy.

\bibliography{GRNN}

\end{document}